\documentclass[letterpaper, 10 pt, conference]{ieeeconf}  

\IEEEoverridecommandlockouts                              

\usepackage{graphics} 
\usepackage{epsfig} 
\usepackage{caption}
\usepackage{mathptmx} 
\usepackage{times} 
\usepackage{amsmath} 
\usepackage{amssymb}  
\usepackage{algorithm}
\usepackage{algpseudocode}
\usepackage{multirow}
\usepackage{booktabs} 
\usepackage[most]{tcolorbox} 
\usepackage{xcolor}
\usepackage{newunicodechar}
\usepackage{caption}
\usepackage[style=ieee]{biblatex} 
\usepackage{hyperref}
\hypersetup{
    hypertex=true,
    colorlinks=true,
    linkcolor=red,     
    anchorcolor=red,   
    citecolor=black,    
    urlcolor=blue      
}

\newunicodechar{（}{(} 
\newunicodechar{）}{)} 

\title{\LARGE \bf
Manipulation Facing Threats: Evaluating Physical Vulnerabilities in End-to-End Vision Language Action Models
}

\author{
Hao Cheng$^{1, 4, 8*}$, Erjia Xiao$^{1*}$, Yichi Wang$^{6*}$, Chengyuan Yu$^7$, Mengshu Sun$^{6}$, Qiang Zhang$^{1,8}$, Jiahang Cao$^{1,8}$,\\
Yijie Guo$^{8}$, Ning Liu$^{8}$, Kaidi Xu$^{5}$, Jize Zhang$^{4}$, Chao Shen$^{3}$, Philip Torr$^{2}$, Jindong Gu$^{2\dagger}$, Renjing Xu$^{1\dagger
}$  \\
{\small $^1$The Hong Kong University of Science and Technology (Guangzhou); 
$^2$University of Oxford; $^3$Xi'an Jiaotong University;
} \\ 
    {\small $^4$ The Hong Kong University of Science and Technology;  $^5$City University of Hong Kong;} \\
    {\small $^6$Beijing University of Technology; $^7$ Duke University; $^8$ X-Humanoid} \\
 {\tt\scriptsize Project Page: \url{https://chaducheng.github.io/Manipulate-Facing-Threats/}}
\thanks{$^*$Equal contribution; 
$^{\dagger}$Corresponding authors.}
\\
}

\begin{document}

\maketitle
\thispagestyle{empty}
\pagestyle{empty}

\begin{abstract}

Recently, driven by advancements in Multimodal Large Language Models (MLLMs), Vision Language Action Models (VLAMs) are being proposed to achieve better performance in open-vocabulary scenarios for robotic manipulation tasks. Since manipulation tasks involve direct interaction with the physical world, ensuring robustness and safety during the execution of this task is always a very critical issue. In this paper, by synthesizing current safety research on MLLMs and the specific application scenarios of the manipulation task in the physical world, we comprehensively evaluate VLAMs in the face of potential physical threats. Specifically, we propose the Physical Vulnerability Evaluating Pipeline  (PVEP) that can incorporate as many visual modal physical threats as possible for evaluating the physical robustness of VLAMs. The physical threats in PVEP specifically include Out-of-Distribution, Typography-based Visual Prompt, and Adversarial Patch Attacks. By comparing the performance fluctuations of VLAMs before and after being attacked, we provide generalizable \textbf{\textit{Analyses}} of how VLAMs respond to different physical threats. 


\end{abstract}

\section{Introduction}

As a task with widespread applications in real-life and industrial manufacturing~\cite{negrello2020hands, ghodsian2023mobile, li2022robotic, davidson2020robotic, wu2023tidybot, yang2023transferring, wu2024new, yoshikawa2023large}, the continuous performance improvements of robotic arm manipulation systems in recent years have been driven by the development of various Artificial Intelligence (AI) algorithms.  
Previously, research on AI-driven manipulation systems predominantly focused on training-from-scratch imitation learning methods, such as behavior cloning~\cite{florence2022implicit, wang2023identifying} and diffusion policy~\cite{ma2024hierarchical, xian2023chaineddiffuser, ze20243d}. 
However, with the emergence of Large Language Models (LLMs)~\cite{alayrac2022flamingo, touvron2023llama} as well as Multimodal Large Language Models (MLLMs)~\cite{zhu2023minigpt, liu2023llava, liu2023improvedllava}, LLMs/MLLMs-driven manipulation systems have also been introduced.

\begin{figure}[t!]
	\setlength{\tabcolsep}{1.0pt}
	\centering
	\begin{tabular}{c}
		\includegraphics[width=0.48\textwidth]{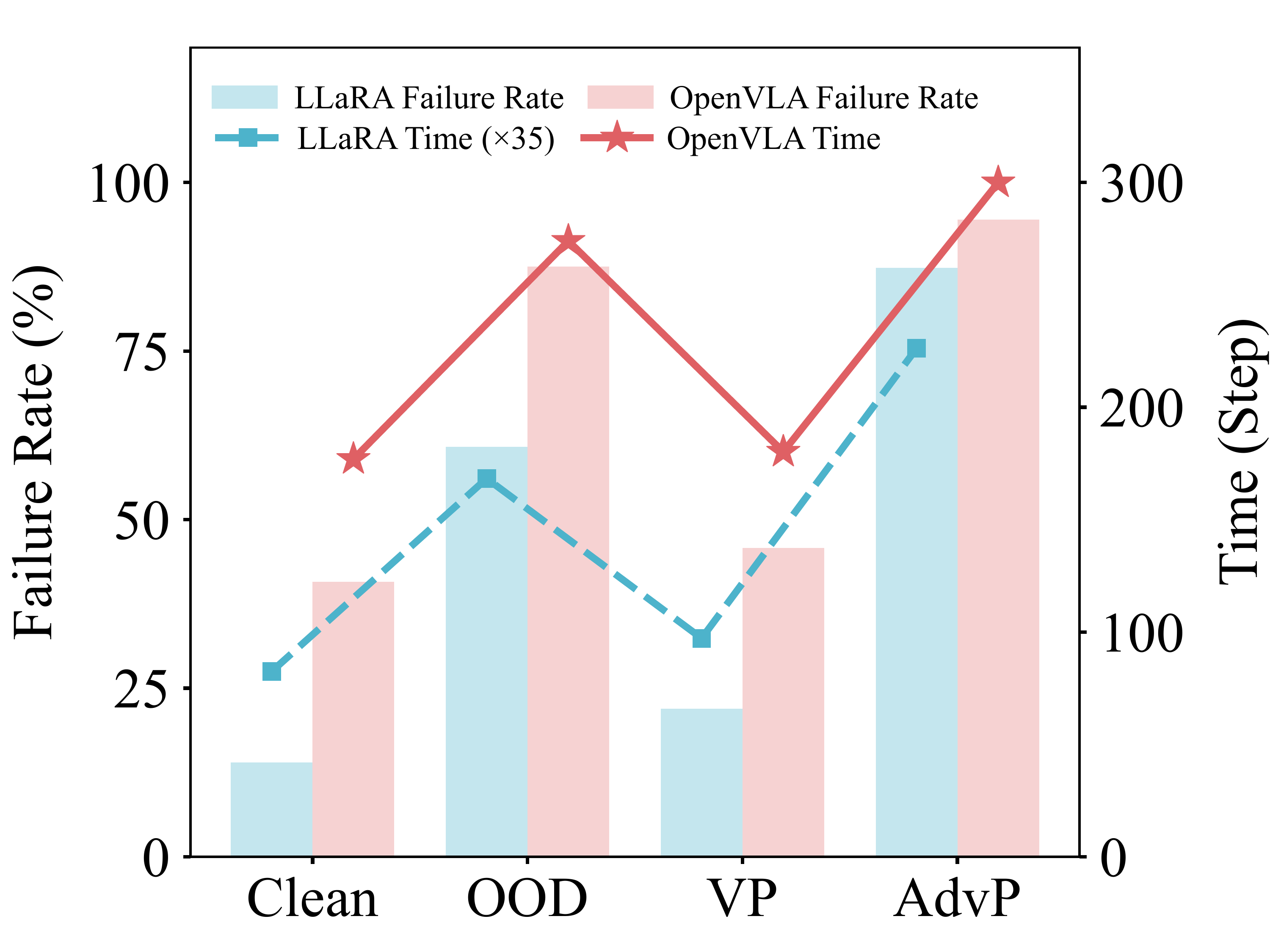} \\
		
	\end{tabular}
        \caption{\small{Performance degradation and time delay of LLaRA and OpenVLA due to physical attacks ($\times 35$ is for better illustration).}}
	\label{fig:good-perfom}
	\vspace{-8mm}
\end{figure}

Compared to imitation learning models trained on single tasks, LLMs/MLLMs-driven manipulation systems 
could gain better performance in open-vocabulary scenarios because of the powerful informative capacity of large models.
Among these, systems that directly leverage commercial closed-source API to enhance manipulation task performance in open-vocabulary situations have been applied in various areas~\cite{wu2023tidybot, yang2023transferring, wu2024new, yoshikawa2023large}.
However, due to the inconvenience of using commercial closed-source models and the challenges of specific deployment, as well as the goal of achieving real AI democratization, it is crucial to propose the LLMs/MLLMs-driven manipulation systems that allows all researchers to access complete information. 
Based on this, manipulation systems leveraging open-source LLMs have also been proposed.
However, unlike the simultaneous improvement in visual perception and instruction semantic understanding brought by incorporating commercial large model API,
open-source LLMs-driven systems still rely on traditional vision models, such as YOLO~\cite{redmon2016you} and Mask R-CNN~\cite{he2017mask}, for the visual perception module to acquire visual modality information. 
Traditional vision models are typically designed for specific, single-vision tasks, which result in a lack of generality and scalability when applied to different zero-shot tasks.
During the training process, these models learn from manually annotated, simple vision-label information pairs, which limits their ability to develop a more comprehensive and complex understanding of visual information when combined with rich language descriptions.
This would incur the following limitations for LLMs-driven manipulation systems in open-vocabulary scenarios: (1) inability to perceive and handle unseen or zero-shot novel objects and environments; (2) inability to understand complex semantic instructions, which prevents the system from responding to more sophisticated tasks. 
To overcome the aforementioned limits, building upon the design principles of MLLMs, the end-to-end Vision-Language-Action Models (VLAMs) have been proposed. 

\begin{figure*}[h!]
\centering
\includegraphics[width=1.0\linewidth]{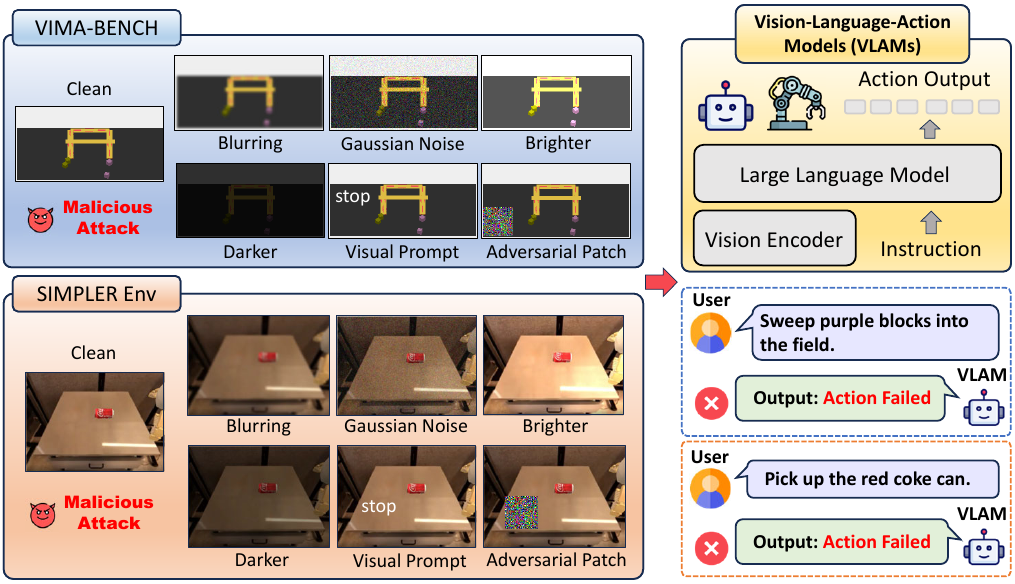}
\caption{\small {The framework for evaluating VLAMs utilizing Physical Vulnerability Evaluation Pipeline (PVEP).}}
\label{framework}
\end{figure*}

By leveraging the vision encoder of MLLMs with powerful visual perception capabilities to obtain high-level visual modality information,
the performance of VLAMs could be further improved in more vocabulary scenarios. 
Moreover, unlike LLMs-driven manipulation systems that require multiple modules to generate policy codes for indirectly manipulating a robotic arm, VLAMs can directly respond to visual modality information based on the given language instructions, enabling end-to-end generation of the corresponding action output for direct manipulation.
Based on these advantages have become a research hotspot in AI-driven robotic manipulation systems. 


Additionally, since robotic manipulation tasks are executed directly in the real world, performance stability in the face of physical security threats becomes critically important.
This leads us to pose an unignorable question:

\textbf{\textit{How safe are AI-driven manipulation systems?}}


As for the physical security threats to AI-driven manipulation systems, \cite{tsurumine2022goal} and \cite{chen2024diffusion} investigate the potential safety issues associated with using imitation learning. 
\cite{wu2024safety, li2024mmro, zhang2024badrobot} explore how
commercial API and open-source LLMs-driven manipulation systems experience performance degradation due to hallucination and jailbreak issues.
As for the MLLMs-driven end-to-end VLAMs, no related physical security validation work has been introduced so far.
Therefore, this paper conducts a comprehensive robustness and safety evaluation of open-source VLAMs, represented by LLaRA~\cite{li2024llara} and OpenVLA~\cite{kim2024openvla}.
For the specific evaluation approaches that may arise when applying VLAMs under fully open-vocabulary scenarios within the physical world. 
We do not include attack methods targeting the language instructions in our threat considerations.
The specific reason is that in physical attack scenarios, VLAMs mainly face relatively unchanged visual modality information, along with various semantic and linguistic template language instructions to process.
Therefore, compared to constantly modifying language modality input, attacks targeting the visual modality input would have a more profound impact on the physical world. 
Based on this, we propose the Physical Vulnerability Evaluation Pipeline (PVEP) to assess the performance variations of VLAMs when confronting visual modality safety threats. 
PVEP includes most currently possible visual attacks in the physical world, such as Out-of-Distribution (OOD), Typography-based Visual Prompt (VP), and Adversarial Patch (AdvP) Attacks, to the best of our knowledge. 
Measured by the failure rate and the number of timesteps to complete tasks,
Figure \ref{fig:good-perfom} shows the largest performance degradation and corresponding time delay of VLAMs before (Clean) and after suffering various physical attacks (OOD, VP, AdvP). Furthermore,
through comprehensive evaluations in PVEP, we present four generalizable \textbf{\textit{Analyses}} for VLAMs in open-vocabulary scenarios:

\textbf{\textit{Analyses: 1. The types and intensities of OOD influence the severity of the attack; 2. The impact of typography-based visual prompt on final output is dependent on the specific VLAMs types and the semantics of typographic text; 3. VLAMs are susceptible to adversarial patches that can affect their final output in the physical world. 4. Since current VLAMs are fine-tuned from MLLMs, adversarial patches generated by MLLMs handling visual tasks exhibit adversarial transferability to VLAMs performing robotic tasks. However, the strength of this transferability depends on the specific type of MLLMs used.}}

Our main contribution is as follows:
\begin{itemize}
    \item We propose a Physical Vulnerability Evaluating Pipeline (PVEP) that allows for the evaluation of visual modality physical security for all existing and future VLAMs.
    \item Based on PVEP, we conduct the most comprehensive robust performance evaluation to date for cutting-edge open-source VLAMs under physical threats. 
    \item Based on our experimental results, we propose four generalizable \textbf{\textit{Analyses}} of performing various physical threats in VLAMs.
\end{itemize}



\section{Background}

\textbf{LLMs/MLLMs-driven Manipulation systems:}
The recent emergence of commercial closed-source large-model APIs, represented by GPT-4~\cite{achiam2023gpt} and Claude~\cite{claude}, marks a significant step forward in the research toward Artificial General Intelligence (AGI).
In contrast to closed-source models, numerous open-source LLMs and MLLMs are proposed to promote the democratization of AI development. For open-source LLMs, Flamingo~\cite{alayrac2022flamingo} LLama~\cite{touvron2023llama}, Vicuna~\cite{kassem2024alpaca} and others~\cite{yao2023llm, rawte2023survey, yin2023survey} obtain good performance in various zero-shot natural language processing tasks. 
Open-source MLLMs, such as LLaVA, MiniGPT-4 and others~ \cite{zhu2023minigpt, liu2023llava, liu2023improvedllava, karamcheti2024prismatic}, could be obtained through fusing open-source pre-trained vision encoders~\cite{radford2021learning} with different open-source LLMs.
Based on this progress, there are lots of commercial API-driven manipulation systems have been adopted widespread applications in different areas, such as home-services robot~\cite{wu2023tidybot, yang2023transferring}, agriculture robot~\cite{wu2024new}, scientific experiments robot~\cite{yoshikawa2023large}.
Then, referring to the advancement of open-source LLMs, some LLMs-driven manipulation systems, such as VIMA and others~\cite{jiang2022vima, nair2022r3m}, are also proposed.
Additionally, about MLLMs-driven manipulation systems, a promising trend has emerged in the form of Vision-Language-Action models (VLAMs), which involve fine-tuning large pre-trained MLLMs for robot action prediction. These models are characterized by their fusion of robot control actions directly into MLLMs backbones.
RT-2~\cite{brohan2023rt} and PaLM~\cite{anil2023palm, driess2023palm} have garnered significant attention due to their claims of achieving promising performance across various manipulation tasks. However, both of them are closed-source.
Therefore, LLaRA~\cite{li2024llara} and OpenVLA~\cite{kim2024openvla} as the totally open-source VLAMs are proposed. 
Since open-source models can be continuously evolved due to their easy accessibility, these two VLAMs may potentially revolutionize autonomous systems and human-robot interaction paradigms in the future.

\textbf{Safety Concerns:}
Different types of physical visual attacks could significantly impact the performance of various AI models.
OOD attacks~\cite{lee2018simple, besnier2021triggering, herrmann2022pyramid, wang2024llms} can consistently affect the stability of AI models.
For the recently emerged LLMs and MLLMs, there are also issues such as jailbreak~\cite{wei2024jailbroken, huang2023catastrophic, wang2024llms, xu2024defending} and hallucination~\cite{yao2023llm, rawte2023survey, tonmoy2024comprehensive, xu2024hallucination} that can undermine the reliability of the final language output.
Besides, for MLLMs, typography-based visual prompts~\cite{azuma2023defense, cheng2024typographic} could distract the semantics of final language output by adding simple pixel-level text to visual modality input.
In addition, various kinds of AI models, including LLMs and MLLMs, as victim models, have been certified to possess adversarial vulnerabilities~\cite{bai2021recent, mao2022towards, kumar2023certifying, wang2024stop}. 
The output of AI models can be altered by adversarial perturbations, leading to either deviation from the correct output (untargeted) or convergence towards a predefined incorrect output (targeted).
Furthermore, the generating process of such perturbations could be classified into black and white box, depending on the amount of information attackers possess about the victim model.
About the particular vulnerability of AI-driven manipulation systems, \cite{tsurumine2022goal} and \cite{chen2024diffusion} investigate the potential safety issues of using diffusion policy during executing manipulation tasks.
\cite{zhang2024badrobot} explores the jailbreak threats in commercial API-driven manipulation systems. And
\cite{li2024mmro} introduces the MMRo benchmark, designed to evaluate  API-driven systems across four key areas: perception, task planning, visual reasoning, and safety.
\cite{wu2024safety} evaluates safety challenges of open-source LLMs-driven manipulation systems in vision-language modality. 
However, there is no physical threat evaluation work targeting open-source VLAMs so far. 
\section{Physical Vulnerability Evaluating Pipeline}
Figure~\ref{framework} illustrates the overall framework for evaluating physical security threats to VLAMs using the Physical Vulnerability Evaluating Pipeline (PVEP).

\subsection{Preliminaries of Physical Attacks}

PVEP includes three of the most common physical visual threat methods in real-world environments: Out-of-Distribution (OOD), visual prompts, and adversarial patch attacks.
$\{x, t\}$ is the vision-language input pairs, $I_{type}(x)$ is different types of physical visual attack methods. 

\textbf{ Out-of-Distribution (OOD):}
For OOD attacks, we specifically use Blurring (Blur), Gaussian Noise (GN), and Brightness Control (BC).

\begin{align}
    & I_{\text{Blur}}(x) = \frac{1}{2\pi\sigma_{Gk}^2} \exp\left(-\frac{x^2}{2\sigma_{Gk}^2}\right)
\end{align}

where $\sigma$ is the standard deviation of the Gaussian kernel to control the strength of the blurring effect. $\frac{1}{2\pi\sigma^2}$ is the normalization item of the Blur operation.

\begin{equation}
\centering
     I_{\text{GN}}(x) = x + N(x) \quad s.t. N(x) \sim \mathcal{N}(\mu, \sigma_{Gn}^2)
\end{equation}

where $N(x) \sim \mathcal{N}(\mu, \sigma^2)$ is the Gaussian noise. $\mu$ is the mean of the noise, and $\sigma$ is the deviation of the added Gaussian noise.

\begin{align}
    I_{\text{BC}}(x) = x \times \alpha
\end{align}

$\alpha \in [0.0, 2.0]$ is the brightness factor.
When $\alpha>1$ and $\alpha < 1$, the image will be Brighter (BCB) and Darker (BCD)

\textbf{Typography-based Visual Prompts:}

\begin{equation}
\centering
     I_{\text{Typo}}(x) = x + t
\end{equation}

$t$ is the typographic text with different semantics that could be directly added to the original images.

\textbf{Physical Adversarial Patch}





\begin{align}
I_{\text{adv}}(x) = \min_{\delta \in S} L(f(\theta, x \odot (1-m) + \delta \odot m), y_t)
\end{align}

$f(\theta, x, t)$ is the victim model, $\delta$ is the adversarial patch, $y_t$ is the targeted output, $S$ is the constraint set for $\delta$, $m$ is the binary mask indicating the patch location, $\odot$ denotes element-wise multiplication

\subsection{Threat Models}

When adopting VLAMs, \textbf{\textit{Users}} first encounter a specific manipulation visual scene. Then, by applying language instructions with the fixed template, VLAMs generate multiple visual-instruction information pairs corresponding to the sequential action steps required to execute the manipulation task. 
In this process, it is explicitly stated that \textbf{\textit{Users}} can only access VLAMs at the black-box level.
Therefore, as \textbf{\textit{Attackers}}, when the amount of information available is just equivalent to that of a regular user, this constitutes a black-box attack. Whereas attackers have access to all the structural and parameter information of VLAMs, this type of attack is considered as the white-box attack.
Specifically, for each type of physical attack, both OOD and visual prompts are black-box level attacks because they only perform pixel-level editing on the visual modality information. 
For the generation process of adversarial patches, there are both black-box and white-box approaches.
The white-box adversarial patch generation process involves accessing the combination of all available information of VLAMs and visual-instruction information pairs. 
For black-box adversarial patch generation, \textbf{\textit{Attackers}} can leverage the adversarial transfer characteristics across different AI models.
All currently available open-source VLAMs are fine-tuned based on certain open-source MLLMs, and these MLLMs could be easily accessed in full detail due to the availability of numerous pretrained models online for download.
It is easily achievable to generate a corresponding physical adversarial patch based on the information of MLLMs and apply this patch to VLAMs to execute the transferable attack. 

Additionally, it is crucial to analyze the impact of specific types of input image-language modality pairs on the adversarial transferability performance of VLAMs.
Algorihtm~\ref{Algorithm:transfer} presents the execution process of the adversarial transfer attack from MLLMs to VLAMs.
For black-box \textbf{\textit{Attackers}}, compared to robotic vision-instruction language pairs $\{x_{rv}, t_{vi}\}$ for VLAMs, which typically requires simulation or real-world scene recording to obtain, the general VQA image-prompt pairs $\{x_{gi}, t_{gp}\}$ of MLLMs are often more easily available as the online dataset, such as DAQUAR \cite{malinowski2014multi}, TallyQA\cite{acharya2019tallyqa}, A-OKVQA \cite{schwenk2022okvqa} and others.


\begin{algorithm}
\begin{algorithmic}[1]
\State \textbf{Input:} 
Vision and language pairs input $\{x_i, t_j\}$, where $\{i,j\}$ could be general VQA image-prompt pairs $\{gi, gp\}$ or robotic vision-instruction information pairs $\{rv, ri\}$; MLLMs $f_m({\theta}_m, x, t)$; VLAMs $f_v({\theta}_v, x, t)$; targeted output $y_t$; mask $m$.
\State \textbf{Output:} Adversarial patch $\delta$.
\State Initialize $\delta$ randomly within constraints $S$
\State $\delta \gets \min_ L(f_m({\theta}_m, x_i \odot (1-m) + \delta \odot m, t_j), y_L)$  
\Comment{$\{x_i, t_j\}$ could be $\{x_{gi}, t_{gp}\}$ or $\{x_{rv}, t_{ri}\}$}
\State $y_t \gets f_v({\theta}_v, x_{rv} \odot (1-m) + \delta \odot m, t_{ri})$ 
\Comment{Apply $\delta$ generated from MLLMs to VLAMs}
\State \Return $\delta$
\end{algorithmic}
\caption{Transferable Attack from MLLMs to VLAMs}
\label{Algorithm:transfer}
\end{algorithm}
\section{Experiments}

\begin{table*}[t!]
\centering
\captionsetup{justification=centering, singlelinecheck=false}
\caption{\small {Failure rates (\%) of LLaRA on 14 VIMA tasks under 3 physical attack categories ({\color[HTML]{FF0000} Red} is {\color[HTML]{FF0000}$\uparrow$}, {\color[HTML]{009901} Green} is {\color[HTML]{009901}$\downarrow$})}} 
\scriptsize
\scalebox{0.8}{
\setlength{\tabcolsep}{1mm}{
\begin{tabular}{c|c|cccc|ccccccc|cccc}
\toprule[1.2pt]
&                         & \multicolumn{4}{c|}{\textbf{Out-of-Distribution}}                                                                                                                          & \multicolumn{7}{c|}{\textbf{Typography-based Visual Prompt}}                                                                                                                                                                                                                             & \multicolumn{4}{c}{\textbf{Adversarial Patch}}                                                                                                                      \\ \cline{3-17} 
\multirow{-2}{*}{\begin{tabular}[c]{@{}c@{}}\textbf{Failure}\\ \textbf{Rate (\%)} \end{tabular}} & \multirow{-2}{*}{\textbf{Clean}} & \textit{Blur}                               & \textit{GN}                                 & \textit{BC(B)}                              & \textit{BC(D)}                              & \textit{TW1}                                & \textit{TW2}                               & \textit{TW3}                                & \textit{TW4}                                & \textit{TN1}                                & \textit{TN2}                                & \textit{TN3}                                & \textit{BB}                                 & \textit{RBB}                                & \textit{GB}                                 & \textit{WB}                                  \\ \midrule[1.2pt]
\textbf{\textit{LT1}}                                                                           & 7.5                     &  75.0 {\color[HTML]{FF0000}(67.5)} & 50.0 {\color[HTML]{FF0000} (42.5)} & 10.0 {\color[HTML]{FF0000}  (2.5)}                         & 15.0 {\color[HTML]{FF0000} (7.5)}  & 10.0 {\color[HTML]{FF0000}  (2.5)}  & 2.5 {\color[HTML]{009901}  (5.0)}  & 10.0 {\color[HTML]{FF0000}  (2.5)}  & 12.5 {\color[HTML]{FE0000}  ($\uparrow$5.0)}  &5.0  {\color[HTML]{009901} (2.5)}   & 5.0 {\color[HTML]{009901}  (2.5)}   &5.0  {\color[HTML]{009901} (2.5)}   & 9.8 {\color[HTML]{FE0000}  (2.3)}   & 9.8 {\color[HTML]{FE0000}  (2.3)}   & 10.8 {\color[HTML]{FE0000}  (3.3)}  &  87.5 {\color[HTML]{FE0000} (80.0)}  \\
\textbf{\textit{LT2}}                                                                          & 8.3                     & 26.7{\color[HTML]{FF0000}  (18.4)} & 23.3 {\color[HTML]{FF0000}  (15.0)} & 1.7 {\color[HTML]{009901}  (6.6)}   & 3.3 {\color[HTML]{009901}  (5.0)}   & 10.0 {\color[HTML]{FF0000}  (1.7)}  & 8.3 (0.0)                         & 8.3 (0.0)                          & 5.0 {\color[HTML]{009901} (3.3)}   & 5.0 {\color[HTML]{009901} 5.0 (3.33)}  & 5.0 {\color[HTML]{009901} (3.3)}   & 6.7 {\color[HTML]{009901} (1.6)}   & 6.7 {\color[HTML]{009901} (1.6)}   & 11.5 {\color[HTML]{FE0000} (3.2)}  & 10.0 {\color[HTML]{FE0000} (1.7)}  & 98.3 {\color[HTML]{FE0000} (90.0)}  \\
\textbf{\textit{LT3}}                                                                           & 5.0                     & 36.7 {\color[HTML]{FF0000} (31.7)} & 35.0 {\color[HTML]{FF0000} (30.0)} & 0.0 {\color[HTML]{009901} (5.0)}   & 5.0 (0.0)                          & 6.7 {\color[HTML]{FF0000} (1.7)}   & 3.3 {\color[HTML]{009901} (1.7)}  & 8.3 {\color[HTML]{FE0000} (3.3)}   & 8.3 {\color[HTML]{FE0000} (3.3)}   & 5.0 (0.0)                          & 3.3 {\color[HTML]{009901} (1.7)}   & 5.0 (0.0)                          & 8.0 {\color[HTML]{FE0000} (3.0)}   & 9.8 {\color[HTML]{FE0000} (4.8)}   & 10.7 {\color[HTML]{FE0000} (5.7)}  & 100.0 {\color[HTML]{FE0000} (95.0)} \\
\textbf{\textit{LT4}}                                                                           & 1.7                     & 45.0 {\color[HTML]{FF0000} (43.3)} & 16.7 {\color[HTML]{FF0000} (15.0)} & 1.7 (0.0)                          & 1.7 (0.0)                          & 1.7 (0.0)                          & 1.7 (0.0)                         & 1.7 (0.0)                          & 0.0 {\color[HTML]{009901} (1.7)}   & 1.7 (0.0)                          & 1.7 (0.0)                          & 1.7 (0.0)                          & 0.8 {\color[HTML]{009901} (0.9)}   & 1.8 {\color[HTML]{FE0000} (0.1)}   & 1.8 {\color[HTML]{FE0000} (0.1)}   & 98.3 {\color[HTML]{FE0000} (96.6)}  \\
\textbf{\textit{LT5}}                                                                           & 10.0                    & 60.0 {\color[HTML]{FF0000} (50.0)} & 61.7 {\color[HTML]{FF0000} (51.7)} & 11.7 {\color[HTML]{FF0000} (1.7)}  & 25.0 {\color[HTML]{FF0000} (15.0)} & 13.3 {\color[HTML]{FE0000} (3.3)}  & 10.0 (0.0)                        & 25.0 {\color[HTML]{FE0000} (15.0)} & 23.3 {\color[HTML]{FE0000} (13.3)} & 10.0 (0.0)                         & 15.0 {\color[HTML]{FE0000} (5.0)}  & 15.0 {\color[HTML]{FE0000} (5.0)}  & 12.7 {\color[HTML]{FE0000} (2.7)}  & 20.8 {\color[HTML]{FE0000} (10.8)} & 15.5 {\color[HTML]{FE0000} (5.5)}  & 98.3 {\color[HTML]{FE0000} (88.3)}  \\
\textbf{\textit{LT6}}                                                                          & 18.3                    & 70.0 {\color[HTML]{FF0000} (51.7)} & 35.0 {\color[HTML]{FF0000} (16.7)} & 16.7 {\color[HTML]{009901} (1.6)}  & 20.0 {\color[HTML]{FF0000} (1.7)}  & 15.0 {\color[HTML]{009901} (3.3)}  & 18.3 (0.0)                        & 21.7 {\color[HTML]{FE0000} (3.4)}  & 23.3 {\color[HTML]{FE0000} (5.0)}  & 16.7 {\color[HTML]{009901} (1.6)}  & 18.3 (0.0)                         & 15.0 {\color[HTML]{009901} (3.3)}  & 15.7 {\color[HTML]{009901} (2.6)}  & 19.5 {\color[HTML]{FE0000} (1.2)}  & 19.0 {\color[HTML]{FE0000} (0.7)}  & 98.3 {\color[HTML]{FE0000} (80.0)}  \\
\textbf{\textit{LT7}}                                                                           & 6.7                     & 10.0 {\color[HTML]{FF0000} (3.3)}  & 13.3 {\color[HTML]{FF0000} (6.6)}  & 5.0 {\color[HTML]{009901} (1.7)}   & 1.7 {\color[HTML]{009901} (5.0)}   & 3.3 {\color[HTML]{009901} (3.4)}   & 10.0 {\color[HTML]{FE0000} (3.3)} & 3.3 {\color[HTML]{009901} (3.4)}   & 11.7 {\color[HTML]{FE0000} (5.0)}  & 8.3 {\color[HTML]{FE0000} (1.6)}   & 6.7 (0.0)                          & 3.3 {\color[HTML]{009901} (3.4)}   & 3.8 {\color[HTML]{009901} (2.9)}   & 5.8 {\color[HTML]{009901} (0.9)}   & 4.2 {\color[HTML]{009901} (2.5)}   & 100.0 {\color[HTML]{FE0000}  (93.3)} \\
\textbf{\textit{LT8}}                                                                           & 5.0                     & 71.7 {\color[HTML]{FF0000} (66.7)} & 33.3 {\color[HTML]{FF0000} (28.3)} & 11.7 {\color[HTML]{FE0000} (6.7)}  & 15.0 {\color[HTML]{FE0000} (10.0)} & 11.7 {\color[HTML]{FF0000} (6.7)}  & 3.3 {\color[HTML]{009901} (1.7)}  & 15.0 {\color[HTML]{FE0000} (10.0)} & 15.0 {\color[HTML]{FE0000} (10.0)} & 6.7 {\color[HTML]{FE0000} (1.7)}   & 8.3 {\color[HTML]{FE0000} (3.3)}   & 6.7 {\color[HTML]{FE0000} (1.7)}   & 10.0 {\color[HTML]{FE0000} (5.0)}  & 12.3 {\color[HTML]{FE0000} (7.3)}  & 9.5 {\color[HTML]{FE0000} (4.5)}   & 80.0 {\color[HTML]{FE0000} (75.0)}  \\
\textbf{\textit{LT9}}                                                                          & 6.7                     & 50.0 {\color[HTML]{FF0000} (43.3)} & 43.3 {\color[HTML]{FF0000} (36.6)} & 13.3 {\color[HTML]{FE0000} (6.6)}  & 16.7 {\color[HTML]{FE0000} (10.0)} & 11.7 {\color[HTML]{FF0000} (5.0)}  & 5.0 {\color[HTML]{009901} (1.7)}  & 11.7 {\color[HTML]{FE0000} (5.0)}  & 20.0 {\color[HTML]{FE0000} (13.3)} & 6.7 (0.0)                          & 6.7 (0.0)                          & 10.0 {\color[HTML]{FE0000} (3.3)}  & 8.7 {\color[HTML]{FE0000} (2.0)}   & 14.3 {\color[HTML]{FE0000} (7.6)}  & 10.8 {\color[HTML]{FE0000} (4.1)}  & 100.0 {\color[HTML]{FE0000} (93.3)} \\
\textbf{\textit{LT10}}                                                                         & 5.0                     & 78.3 {\color[HTML]{FF0000} (73.3)} & 51.7 {\color[HTML]{FF0000} (46.7)} & 15.0 {\color[HTML]{FE0000} (10.0)} & 15.0 {\color[HTML]{FE0000} (10.0)} & 5.0 (5.0)                          & 10.0 {\color[HTML]{FF0000} (5.0)} & 13.3 {\color[HTML]{FE0000} (8.3)}  & 13.3 {\color[HTML]{FE0000} (8.3)}  & 11.7 {\color[HTML]{FE0000} (6.7)}  & 11.7 {\color[HTML]{FE0000} (6.7)}  & 8.3 {\color[HTML]{FE0000} (3.3)}   & 16.2 {\color[HTML]{FE0000} (11.2)} & 20.8 {\color[HTML]{FE0000} (15.8)} & 20.0 {\color[HTML]{FE0000} (15.0)} & 63.3 {\color[HTML]{FE0000} (58.3)}  \\
\textbf{\textit{LT11}}                                                                         & 11.7                    & 70.0 {\color[HTML]{FF0000} (58.3)} &  46.7 {\color[HTML]{FF0000}(35.0)} & 15.0 {\color[HTML]{FE0000} (3.3)}  & 28.3 {\color[HTML]{FE0000} (16.6)} & 21.7 {\color[HTML]{FE0000} (10.0)} & 13.3 {\color[HTML]{FF0000} (1.6)} & 23.3 {\color[HTML]{FE0000} (11.6)} & 26.7 {\color[HTML]{FE0000} (15.0)} & 11.7 (0.0)                         & 20.0 {\color[HTML]{FE0000} (8.3)}  & 16.7 {\color[HTML]{FE0000} (5.0)}  & 18.0 {\color[HTML]{FE0000} (6.3)}  & 21.3 {\color[HTML]{FE0000} (9.6)}  & 19.8 {\color[HTML]{FE0000} (8.1)}  & 98.3 {\color[HTML]{FE0000} (86.6)}  \\
\textbf{\textit{LT12}}                                                                         & 15.0                    & 83.3 {\color[HTML]{FF0000} (68.3)} & 43.3 {\color[HTML]{FF0000} (28.3)} & 23.3 {\color[HTML]{FE0000} (8.3)}  & 20.0 {\color[HTML]{FE0000} (5.0)}  & 20.0 {\color[HTML]{FE0000} (5.0)}  & 16.7 {\color[HTML]{FF0000} (1.7)} & 26.7 {\color[HTML]{FE0000} (11.7)} & 28.3 {\color[HTML]{FE0000} (13.3)} & 18.3 {\color[HTML]{FE0000} (3.3)}  & 18.3 {\color[HTML]{FE0000} (3.3)}  & 20.0 {\color[HTML]{FE0000} (5.0)}  & 18.7 {\color[HTML]{FE0000} (3.7)}  & 25.0 {\color[HTML]{FE0000} (10.0)} & 26.0 {\color[HTML]{FE0000} (11.0)} & 100.0 {\color[HTML]{FE0000} (85.0)}   \\
\textbf{\textit{LT13}}                                                                         & 40.0                    & 85.0 {\color[HTML]{FF0000} (45.0)} & 80.0 {\color[HTML]{FF0000} (40.0)} & 70.0 {\color[HTML]{FE0000} (30.0)} & 65.0 {\color[HTML]{FE0000} (25.0)} & 50.0 {\color[HTML]{FE0000} (10.0)} & 45.0 {\color[HTML]{FF0000} (5.0)} & 45.0 {\color[HTML]{FE0000} (5.0)}  & 50.0 {\color[HTML]{FE0000} (10.0)}   & 45.0 {\color[HTML]{FE0000}  (5.0)}  & 50.0 {\color[HTML]{FE0000} (10.0)} & 50.0 {\color[HTML]{FE0000} (10.0)} & 51.5 {\color[HTML]{FE0000} (11.5)} & 24.5 {\color[HTML]{FE0000} (14.5)} & 59.5 {\color[HTML]{FE0000} (19.5)} & 100.0 {\color[HTML]{FE0000} (60.0)}   \\
\textbf{\textit{LT14}}                                                                          & 55.0                    & 90.0 {\color[HTML]{FF0000} (35.0)} & 80.0 {\color[HTML]{FF0000} (25.0)} & 65.0 {\color[HTML]{FE0000} (10.0)} & 65.0 {\color[HTML]{FE0000} (10.0)} & 55.0 (0.0)                         & 55.0 (0.0)                        & 75.0 {\color[HTML]{FE0000} (20.0)} & 70.0 {\color[HTML]{FE0000} (15.0)} & 45.0 {\color[HTML]{009901} (10.0)} & 35.0 {\color[HTML]{009901} (20.0)} & 55.0 (0.0)                         & 63.0 {\color[HTML]{FE0000} (8.0)}  & 73.0 {\color[HTML]{FE0000} (18.0)} & 69.5 {\color[HTML]{FE0000} (14.5)} & 100.0 {\color[HTML]{FE0000} (45.0)}   \\ \midrule[1.2pt]
\textbf{\textit{Avg}}                                                                           & 14.0                    & 60.8 {\color[HTML]{FF0000} ($\uparrow$46.8)} & 43.8 {\color[HTML]{FF0000} ($\uparrow$29.8)} & 18.6 {\color[HTML]{FE0000} ($\uparrow$4.6)}  & 21.2 {\color[HTML]{FE0000} ($\uparrow$7.2)}  & 16.8 {\color[HTML]{FF0000} ($\uparrow$2.8)}  & 14.5 {\color[HTML]{FF0000} ($\uparrow$0.5)} & 20.6 {\color[HTML]{FF0000} ($\uparrow$6.6)}  & 22.0 {\color[HTML]{FE0000} ($\uparrow$8.0)}  & 14.1 {\color[HTML]{FE0000} ($\uparrow$0.1)}  & 14.6 {\color[HTML]{FE0000} ($\uparrow$0.6)}  & 15.6 {\color[HTML]{FE0000} ($\uparrow$1.6)}  & 17.4 {\color[HTML]{FE0000} ($\uparrow$3.4)}  & 19.3 {\color[HTML]{FE0000} ($\uparrow$5.3)}  & 20.5 {\color[HTML]{FE0000} ($\uparrow$6.5)}  & 94.5 {\color[HTML]{FE0000} ($\uparrow$80.5)}  \\ 
\bottomrule[1.2pt]
\end{tabular}}}
\label{failure_rate_llara}
\end{table*}

\begin{figure*}[h!]
\centering
\includegraphics[width=1.0\linewidth]{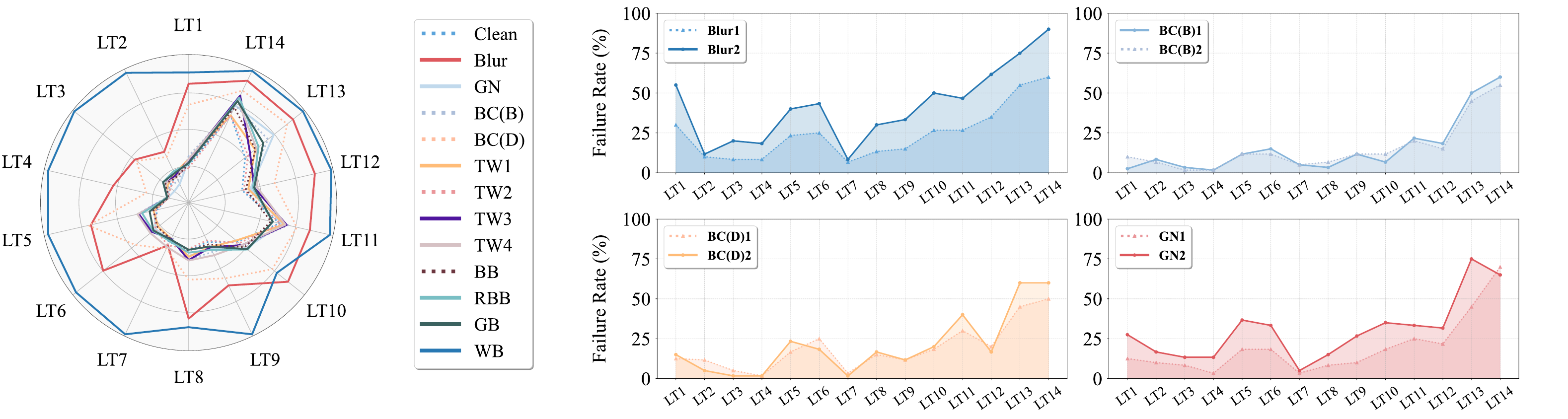}
\caption{ \small {Under 3 physical attack categories: (left) Time steps (with a maximum limit of 8) of LLaRA on 14 VIMA tasks that are listed in TABLE \ref{failure_rate_llara}.
(right) Failure rates of the OOD attacks with other levels that are not listed in TABLE \ref{failure_rate_llara}.}
}
\label{fig_llara}
\vspace{-1mm}
\end{figure*}

\subsection{Models and Simulators}
For VLAMs, we adopt LLaRA~\cite{li2024llara} and OpenVLA~\cite{kim2024openvla} as victim models. The LLaRA and OpenVLA are evaluated on the VIMA~\cite{jiang2022vima} and SimplerEnv~\cite{li2024evaluating} simulator respectively. 

To be more specific, for LLaRA, we utilized the VIMA simulator to test 14 predefined tasks, denoted as LLaRA Task (\textbf{\textit{LT1}} to \textbf{\textit{LT14}}), corresponding to \texttt{\{sweep without exceeding, rotate, scene understanding, visual manipulation, novel adjective, novel noun, follow order, rearrange, manipulate old neighbor, pick in order then restore, rearrange and restore, same shape, novel adjective and noun, follow motion\}}. 
We also evaluated OpenVLA in the SimplerEnv simulator across 6 predefined tasks, denoted as OpenVLA Task (\textbf{\textit{OT1}} to \textbf{\textit{OT6}}), corresponding to \texttt{\{pick coke can, pick horizontal coke can, pick vertical coke can, pick standing coke can, move near v0, move near v1\}}.



\subsection{Physical Attack Settings}

\subsubsection{\textbf{Out-of-Distribution} Attack}

For Out-of-Distribution (OOD) attacks, we implement three methods: Blurring (\textit{Blur}), Gaussian Noise (\textit{GN}), and Brightness Control (\textit{BC}). For each method, we applied four distinct levels of intensity, ranging from mild to severe, to systematically evaluate their impact on model performance.

For the \textit{Blur}, we implement three levels of blurring with increasing radii (2, 4, and 6 pixels), corresponding to progressively stronger blurring effects from mild to severe.
In the \textit{GN}, we establish three levels of noise intensity by varying the variance (0.01, 0.05, and 0.1), while maintaining a constant mean of 0. These levels represent a gradual increase in noise severity, from subtle to pronounced.
As for the Brightness Control attack, we use a multiplicative factor $\alpha$ to adjust image brightness. Specifically, we subdivide the BC attack into two categories, each with four levels of increasing intensity. In BC Brighter (\textit{B}), three levels ($\alpha = 1.2, 1.4, 1.6$) represent a progression from slightly brighter to significantly overexposed images. In BC Darker (\textit{D}), we use three levels ($\alpha = 0.8, 0.4, 0.2$) to represent a progression from slightly darker to severely underexposed images. 

\subsubsection{\textbf{Typography-based Visual Prompt} Attack}

We implement a Typography-based Visual Prompts attack to evaluate the robustness of VLAMs to visual prompt interventions. We categorize our interventions into two distinct types: word types (\textit{TW}) and numerical types (\textit{TN}). This dichotomy allows us to assess whether VLAMs exhibit differential sensitivity to semantic or numerical information when processing visual data. For word types, we select 4 candidate words, denoted as \textit{TW1} to \textit{TW4}, corresponding to \texttt{\{move bottom, move top, move slowly, stop moving\}}. For numerical types, We extract three significant numerical values directly from the VLAMs' specific motion output space, such as coordinates or angles, denoted as \textit{TN1} to \textit{TN3}.

\subsubsection{\textbf{Adversarial Patch} Attack}

We implement the this attack by generating a universal adversarial patch capable of influencing the  victim output across various manipulation tasks, thereby degrading the model's performance in robotics manipulation. As we increase our access to information about the victim VLAMs and the image data they use, we design four levels of attacks: Black Box (\textit{BB}), Robotic Black Box (\textit{RBB}), Gray Box (\textit{GB}), and White Box (\textit{WB}). 

In particular, for the \textit{BB} attack, we utilize minimal information about the victim VLAMs. We employe only the victim VLAMs' base model (pre-fine-tuning) as a surrogate model. For instance, for the victim model LLaRA, we use its base model LLaVA as the surrogate model. The adversarial patch was trained using 5000 images from ImageNet and 200 general Visual Question Answering (VQA) prompts selected from \cite{luo2024image}. For the \textit{RBB} attack, based on BB, we provide additional data information for training the adversarial patch. We collect an additional 2000 Images and 200 prompts of robotics manipulation scenarios when using victim VLAMs for inference. For the \textit{GB} attack, we use the victim VLAMs and the general VQA images/prompts in \textit{BB} for adversarial patch training. As for the \textit{WB} attack, we directly use the victim VLAMs and all available images and prompts from both \textit{BB} and \textit{RBB}. By systematically comparing these four levels of adversarial patch attacks, we can assess the minimum level of model and data information required to generate an effective adversarial patch against VLAMs.

\begin{table*}[t!]
\centering
\captionsetup{justification=centering, singlelinecheck=false}
\caption{ \small {Failure rates (\%) of OpenVLA on 6 SimplerEnv tasks under 3 physical attack categories  ({\color[HTML]{FF0000} Red} is {\color[HTML]{FF0000}$\uparrow$}, {\color[HTML]{009901} Green} is {\color[HTML]{009901}$\downarrow$})
}} 
\scriptsize
\scalebox{0.83}{
\setlength{\tabcolsep}{1mm}{
\begin{tabular}{c|c|cccc|ccccccc|ccc}
\toprule[1.2pt]
  &                             & \multicolumn{4}{c|}{\textbf{Out-of-Distribution}}                                                                                                                           & \multicolumn{7}{c|}{\textbf{Typography-based Visual Prompt}}                                                                                                                                                                                                                              & \multicolumn{3}{c}{\textbf{Adversarial Patch}}                                                                                 \\ \cline{3-16} 
\multirow{-2}{*}{\begin{tabular}[c]{@{}c@{}}\textbf{Failure}\\ \textbf{Rate (\%)} \end{tabular}} & \multirow{-2}{*}{\textbf{Clean}}     & \textit{Blur}                                & \textit{GN}                                 & \textit{BC(B)}                              & \textit{BC(D)}                              & \textit{TW1}                                & \textit{TW2}                                & \textit{TW3}                                & T\textit{W4}                                & \textit{TN1}                                & \textit{TN2}                                & \textit{TN3}                                & \textit{BB}                                 & \textit{RBB}                                & \textit{WB}                                  \\ \midrule[1.2pt]
\textbf{\textit{OT1}}                                                                           & 45.0 & 85.0 {\color[HTML]{FE0000} (40.0)}  & 40.0 {\color[HTML]{009901} (5.0)}  &  45.0 (0.0)  & 25.0 {\color[HTML]{009901} (20.0)} & 50.0 {\color[HTML]{FE0000} (5.0)}  & 30.0 {\color[HTML]{009901} (15.0)} & 45.0 (0.0)                         & 60.0 {\color[HTML]{FE0000} (15.0)} & 55.0 {\color[HTML]{FE0000} (10.0)} & 35.0 {\color[HTML]{009901} (10.0)} & 40.0 {\color[HTML]{009901} (5.0)}  & 45.0 (0.0)                         & 40.0 {\color[HTML]{009901} (5.0)}  & 100.0 {\color[HTML]{FE0000} (55.0)} \\
\textbf{\textit{OT2}}                                                                          & 40.0 & 95.0 {\color[HTML]{FE0000} (55.0)}  & 35.0 {\color[HTML]{009901} (5.0)}  & 35.0 {\color[HTML]{009901} (5.0)}  & 20.0 {\color[HTML]{009901} (20.0)} & 45.0 {\color[HTML]{FE0000} (5.0)}  & 45.0 {\color[HTML]{FE0000} (5.0)}  & 50.0 {\color[HTML]{FE0000} (10.0)} &  40.0 (0.0)  & 35.0 {\color[HTML]{009901} (5.0)}  & 15.0 {\color[HTML]{009901} (25.0)} & 45.0 {\color[HTML]{FE0000} (5.0)}  & 30.0 {\color[HTML]{009901} (10.0)} & 50.0 {\color[HTML]{FE0000} (10.0)} & 100.0 {\color[HTML]{FE0000} (60.0)} \\
\textbf{\textit{OT3}}                                                                           & 65.0                        & 100.0 {\color[HTML]{FE0000} (35.0)} & 70.0 {\color[HTML]{FE0000} (5.0)}  & 55.0 {\color[HTML]{009901} (10.0)} & 60.0 {\color[HTML]{009901} (5.0)} & 60.0 {\color[HTML]{009901} (5.0)} & 65.0  (0.0) & 55.0 {\color[HTML]{009901} (10.0)} & 85.0 {\color[HTML]{FE0000} (20.0)} & 80.0 {\color[HTML]{FE0000} (15.0)} & 85.0 {\color[HTML]{FE0000} (20.0)} & 70.0 {\color[HTML]{FE0000} (5.0)} & 85.0 {\color[HTML]{FE0000} (20.0)} & 70.0 {\color[HTML]{FE0000} (5.0)}  & 100.0 {\color[HTML]{FE0000} (35.0)} \\
\textbf{\textit{OT4}}                                                                           & 55.0 & 90.0 {\color[HTML]{FE0000} (35.0)}  & 25.0 {\color[HTML]{009901} (30.0)} & 45.0 {\color[HTML]{009901} (10.0)} & 20.0 {\color[HTML]{009901} (35.0)} & 25.0 {\color[HTML]{009901} (30.0)} & 40.0 {\color[HTML]{009901} (15.0)} & 30.0 {\color[HTML]{009901} (25.0)} & 35.0 {\color[HTML]{009901} (20.0)} & 25.0 {\color[HTML]{009901} (30.0)} & 25.0 {\color[HTML]{009901} (30.0)} & 35.0 {\color[HTML]{009901} (20.0)} & 5.0 {\color[HTML]{009901} (50.0)}  & 10.0 {\color[HTML]{009901} (45.0)} & 100.0 {\color[HTML]{FE0000} (45.0)} \\
\textbf{\textit{OT5}}                                                                           & 10.0 & 85.0 {\color[HTML]{FE0000} (75.0)}  & 35.0 {\color[HTML]{FE0000} (25.0)} & 55.0 {\color[HTML]{FE0000} (45.0)} & 50.0 {\color[HTML]{FE0000} (40.0)} & 20.0 {\color[HTML]{FE0000} (10.0)} & 30.0 {\color[HTML]{FE0000} (20.0)} & 35.0 {\color[HTML]{FE0000} (25.0)} & 20.0 {\color[HTML]{FE0000} (10.0)} & 25.0 {\color[HTML]{FE0000} (15.0)} & 40.0 {\color[HTML]{FE0000} (30.0)} & 35.0 {\color[HTML]{FE0000} (25.0)} & 50.0 {\color[HTML]{FE0000} (40.0)} & 30.0 {\color[HTML]{FE0000} (20.0)} & 100.0 {\color[HTML]{FE0000} (90.0)} \\
\textbf{\textit{OT6}}                                                                          &  30.0 & 70.0 {\color[HTML]{FE0000} (40.0)}  &  30.0 (0.0)  & 40.0 {\color[HTML]{FE0000} (10.0)} & 40.0 {\color[HTML]{FE0000} (10.0)}  & 25.0 {\color[HTML]{009901} (5.0)}  & 40.0 {\color[HTML]{FE0000} (10.0)} & 20.0 {\color[HTML]{FE0000} (10.0)} & 35.0 {\color[HTML]{FE0000} (5.0)}  & 30.0 (0.0)                         & 30.0 (0.0)                         & 20.0 {\color[HTML]{009901} (10.0)} & 30.0 (0.0)                         & 30.0 (0.0)                         & 100.0 {\color[HTML]{FE0000} (70.0)} \\ \midrule[1.2pt]
\textbf{\textit{Avg}}                                                                          &  40.8 & 87.5 {\color[HTML]{FE0000} ($\uparrow$46.7)}  & 39.2 {\color[HTML]{009901} ($\downarrow$1.6)}  & 45.8 {\color[HTML]{FE0000} ($\uparrow$5.0)}  & 35.8 {\color[HTML]{009901} ($\downarrow$5.0)}   & 37.5 {\color[HTML]{009901} ($\downarrow$3.3)}  & 41.7 {\color[HTML]{FE0000} ($\uparrow$0.9)}  & 39.2 {\color[HTML]{009901} ($\downarrow$1.6)}  & 45.8 {\color[HTML]{FE0000} ($\uparrow$5.0)}  & 41.7 {\color[HTML]{FE0000} ($\uparrow$0.9)}  & 38.3 {\color[HTML]{009901} ($\downarrow$2.5)} & 40.8 (0.0)  &  40.8 (0.0)  & 38.3 {\color[HTML]{009901} ($\downarrow$2.5)}  & 100.0 {\color[HTML]{FE0000} ($\uparrow$59.2)} \\ \bottomrule[1.2pt]
\end{tabular}}}
\label{failure_rate_openvla}
\end{table*}

\begin{figure*}[h!]
\centering
\includegraphics[width=1.0\linewidth]{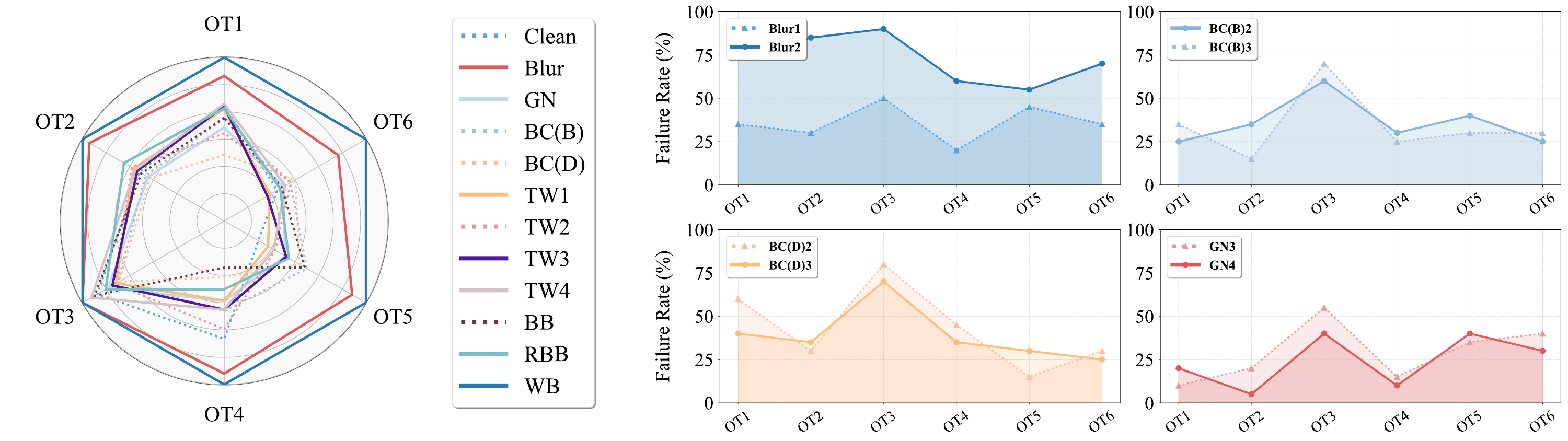}
\caption{\small {Under 3 physical attack categories: (left) Time steps (with a maximum limit of 300) of OpenVLA on 6 SimplerEnv tasks that are listed in TABLE \ref{failure_rate_openvla}. (right) Failure rates of the OOD attacks with other levels that are not listed in TABLE \ref{failure_rate_openvla}.} }
\label{fig_openvla}
\vspace{-1mm}
\end{figure*}

\subsection{Results}

Our comprehensive evaluation of VLAMs under various attack scenarios reveals significant insights into their overall robustness and vulnerabilities. We present different analyses of the average performance (measured by failure rate and number of timesteps to complete a task) across various tasks for 3 physical attack categories.

\subsubsection{\textbf{Out-of-Distribution (Analysis 1)}}

As shown in Table \ref{failure_rate_llara} for LLaRA Tasks, on average, OOD attacks demonstrate varying degrees of effectiveness.
(a) \textit{Blur}: With an average failure rate of 60.8\%, Blur proves to be the most potent OOD attack. This high failure rate suggests that VLAMs are particularly vulnerable to degradation in image clarity.
(b) Gaussian Noise: Showing an average failure rate of 43.8\%, \textit{GN} is less effective than Blur but still poses a significant threat to VLAM performance.
(c) Brightness Control: \textit{BC(B)} yields an average failure rate of 18.6\%. \textit{BC(D)} shows a slightly higher average failure rate of 21.2\%. These results indicate that VLAMs are more robust to brightness changes compared to blurring or noise addition, but still exhibit notable vulnerabilities.

\subsubsection{\textbf{Typography-based Visual Prompt (Analysis 2)}}

Typography attacks reveal interesting patterns in VLAM vulnerabilities to textual and numerical visual prompt interventions. As demonstrated in Table \ref{failure_rate_llara} for LLaRA Tasks, for textual types, \textit{TW1 - TW4} show average failure rates of 16.8\%, 14.5\%, 20.6\%, 22.0\%, respectively. In numerical types, TN1, TN2, and TN3 exhibit average failure rates of 14.1\%, 14.6\%, 15.6\%, respectively. Textual types appear to be slightly more effective than numerical types on average, suggesting that VLAMs might be more sensitive to textual visual prompt interventions in certain contexts.

\subsubsection{\textbf{Adversarial Patch (Analysis 3 \& 4)}}

The adversarial patch attacks reveal a clear escalation in effectiveness as the attacker's knowledge increases. In Table \ref{failure_rate_llara} for LLaRA Tasks, \textit{BB} shows an average failure rate of 17.4\%. \textit{RBB} shows a slightly higher average failure rate of 19.3\%, \textit{RBB} proves marginally more effective than \textit{BB}. \textit{GB} exhibits significantly higher effectiveness, with an average failure rate of 20.5\%. \textit{WB} proves to be the most potent, resulting in an average failure rate of 94.5\%. The dramatic increase in effectiveness from \textit{BB} to \textit{WB} underscores the critical vulnerability of VLAMs when attackers have access to model architecture and training data. \textit{BB} also demonstrates that even with limited information and using only the base model (pre-fine-tuning), attackers can significantly compromise victim VLAM performance.

To further quantify the impact of various physical attack categories on VLAM performance across different tasks, we analyzed the number of timesteps required for task completion. Figure \ref{fig_llara} presents a radar chart illustrating this metric for different LLaRA Tasks under various attack conditions. Consistent with our observations regarding failure rates, the different categories of physical attacks demonstrably increase the number of timesteps needed to complete tasks.
In parallel with our LLaRA experiments, we conducted a similar set of evaluations using OpenVLA on seven distinct tasks within the SimplerEnv simulator. As illustrated in Table \ref{failure_rate_openvla} and Figure \ref{fig_openvla}, the three categories of physical attacks demonstrated comparable effects on OpenVLA's performance. Specifically, these attacks resulted in both increased failure rates and elevated numbers of timesteps required for task completion.

\section{Conclusion}

\textbf{Conclusion:}
This paper proposes the Physical Vulnerability Evaluation Pipeline (PVEP) to comprehensively assess the robustness of Vision-Language-Action Models (VLAMs) against various physical security threats, including Out-of-Distribution, Visual Prompts and Adversarial Patch attacks. 
By conducting detailed performance evaluations of state-of-the-art open-source VLAMs, we propose critical performance summaries of their vulnerability under different real-world physical conditions. Our summaries offer generalizable performance patterns under different threat scenarios, serving as a foundation for future research and development of more robust VLAM systems in robotic manipulation tasks.

\newpage




{\tiny
  \printbibliography  
}

\end{document}